\documentclass[sigconf]{acmart}
\acmSubmissionID{882}
\usepackage{bm}
\usepackage{url}

\AtBeginDocument{%
  \providecommand\BibTeX{{%
    \normalfont B\kern-0.5em{\scshape i\kern-0.25em b}\kern-0.8em\TeX}}}

\copyrightyear{2023} 
\acmYear{2023} 
\setcopyright{rightsretained} 
\acmConference[SIGGRAPH '23 Conference Proceedings]{Special Interest Group
on Computer Graphics and Interactive Techniques Conference Conference
Proceedings}{August 6--10, 2023}{Los Angeles, CA, USA}
\acmBooktitle{Special Interest Group on Computer Graphics and Interactive
Techniques Conference Conference Proceedings (SIGGRAPH '23 Conference
Proceedings), August 6--10, 2023, Los Angeles, CA, USA}
\acmDOI{10.1145/3588432.3591567}
\acmISBN{979-8-4007-0159-7/23/08}

\begin{document}

\title{AvatarMAV: Fast 3D Head Avatar Reconstruction Using Motion-Aware Neural Voxels}

\author{Yuelang Xu}
\affiliation{%
  \institution{Tsinghua University}
  \city{Beijing}
  \country{China}}
\email{xll20@mails.tsinghua.edu.cn}
\orcid{0009-0001-6834-8199}

\author{Lizhen Wang}
\affiliation{%
  \institution{Tsinghua University}
  \institution{NNKosmos Technology}
  \city{Beijing}
  \country{China}}
\email{wlz18@mails.tsinghua.edu.cn}
\orcid{0000-0002-6674-9327}

\author{Xiaochen Zhao}
\affiliation{%
  \institution{Tsinghua University}
  \institution{NNKosmos Technology}
  \city{Beijing}
  \country{China}}
\email{zhaoxc19@mails.tsinghua.edu.cn}
\orcid{0000-0001-8976-7723}

\author{Hongwen Zhang}
\affiliation{%
  \institution{Tsinghua University}
  \city{Beijing}
  \country{China}}
\email{zhanghongwen@mail.tsinghua.edu.cn}
\orcid{0000-0001-8633-4551}

\author{Yebin Liu}
\affiliation{%
  \institution{Tsinghua University}
  \city{Beijing}
  \country{China}}
\email{liuyebin@mail.tsinghua.edu.cn}
\orcid{0000-0003-3215-0225}

\renewcommand{\shortauthors}{Yuelang Xu, et al.}

\begin{abstract}
With NeRF widely used for facial reenactment, recent methods can recover photo-realistic 3D head avatar from just a monocular video. Unfortunately, the training process of the NeRF-based methods is quite time-consuming, as MLP used in the NeRF-based methods is inefficient and requires too many iterations to converge. To overcome this problem, we propose AvatarMAV, a fast 3D head avatar reconstruction method using Motion-Aware Neural Voxels. AvatarMAV is the first to model both the canonical appearance and the decoupled expression motion by neural voxels for head avatar. In particular, the motion-aware neural voxels is generated from the weighted concatenation of multiple 4D tensors. The 4D tensors semantically correspond one-to-one with 3DMM expression basis and share the same weights as 3DMM expression coefficients. Benefiting from our novel representation, the proposed AvatarMAV can recover photo-realistic head avatars in just 5 minutes (implemented with pure PyTorch), which is significantly faster than the state-of-the-art facial reenactment methods. Project page: https://www.liuyebin.com/avatarmav.
\end{abstract}

\ccsdesc[500]{Computing methodologies~Volumetric models}
\keywords{Neural Radiance Field, Facial Reenactment}

\begin{teaserfigure}
  \includegraphics[width=\textwidth]{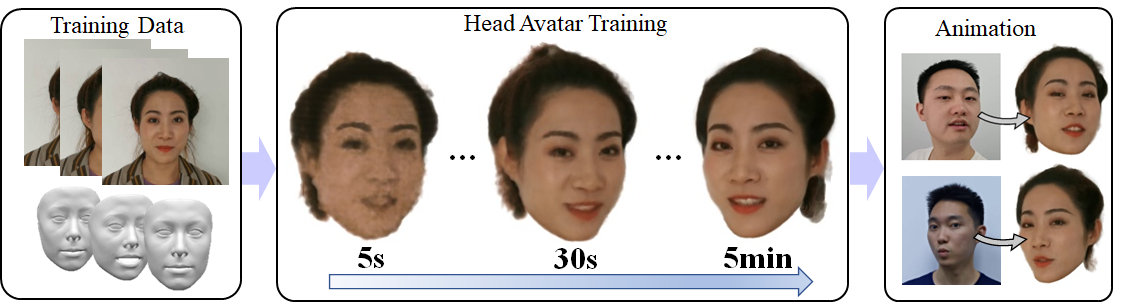}
  \caption{We propose AvatarMAV, a fast 3D head avatar reconstruction method. Given a monocular video, our method can recover photo-realistic head avatar in 5 minutes.}
  \label{fig:teaser}
\end{teaserfigure}

\maketitle

\section{Introduction}

Facial reenactment and head avatar reconstruction from a monocular video have been research hotspots recently, which have a very broad application prospect in VR/AR, digital human, holographic communication, webcast, etc. 

However, current methods cannot reconstruct 3D head avatars in minutes, which has become a main limitation of applications.
Recent works~\cite{gafni2021dynamic, zheng2022imavatar, gao2022reconstructing, grassal2022neural} can recover photo-realistic 3D head avatars using easily available data, such as a monocular video. These methods can be divided into two categories: geometry-based methods and NeRF-based methods. 
Geometry-based~\cite{zheng2022imavatar, grassal2022neural, Khakhulin2022realistic} methods can obtain well-defined 3D face geometry. For example, Neural Head Avatar~\cite{grassal2022neural} constructs the avatar model through non-rigid deformation of the mesh template but is limited by the topology of template itself. IMAvatar~\cite{zheng2022imavatar} solves this problem by optimizing an implicit signed distance field upon a mesh template. These methods model the motion by linear blend skinning of template, which leads to slow training and rendering speed.
NeRF-based methods~\cite{gafni2021dynamic, gao2022reconstructing, guo2021ad, liu2022semantic} achieve photo-realistic and view-consistent rendering and are not limited by the topology and coarse expressiveness of face templates. NeRFace~\cite{gafni2021dynamic} proposes to use an expression conditioned dynamic NeRF to model a head avatar and generates photo-realistic portrait images. However, training a NeRF model often takes hours or even days. The concurrent work on NeRFBlendShape~\cite{gao2022reconstructing} reduces the training time consumption to 20 minutes by introducing the multi-level voxel field representation with multi-resolution hash tables storing features. However, the coupling of motion and appearance still limits their efficiency.



On the other hand, a number of recent approaches~\cite{yu_and_fridovichkeil2021plenoxels, sun2022direct, mueller2022instant} propose to use explicit voxel data structures to represent a static NeRF scene, and their experiments prove that this representation plays a dramatic role in accelerating NeRF training.
Meanwhile, some other methods extend explicit voxel representations to dynamic NeRF. The original D-NeRF~\cite{pumarola2020d} represents dynamic scenes by decoupling deformation field and canonical appearance. TineuVox~\cite{fang2022fast} further accelerates D-NeRF by introducing a voxel grid to model the canonical component. But the deformation field of TineuVox is still built by a deep MLP, which still takes much time to converge.
Based on these methods, it is still not trivial to accelerate dynamic NeRF training in head avatar reconstruction. Specifically, as the human face motions are more complex, a deep MLP to model the complex motions is usually inevitable. As a result, it takes much longer time for the MLP to converge during model training.

To overcome the above challenges, we propose AvatarMAV, a fast 3D head avatar reconstruction method using Motion-Aware Neural Voxels. Our approach can achieve 5-minute 3D head avatar reconstruction which means the training can be completed immediately just after collecting the training data. The key idea consists of two points: First, inspired by previous dynamic NeRF reconstruction approaches\cite{athar2021flame, park2021nerfies, park2021hypernerf}, we decouple the complex expression-related motion from the canonical appearance in NeRF-based head avatar reconstruction. Second, We propose an efficient voxel-based representation instead of deep MLPs, for both the appearance and the motion field. Note that NeRFBlendshape~\cite{gao2022reconstructing} only introduces the voxel-based representation but does not decouple the motion and appearance.
We utilize the prior information provided by 3DMM expression basis to model the expression-related motion using only voxel grids and a tiny MLP. Specifically, we use an expression-conditioned neural voxel grid to describe the motion field, and further decompose this neural voxel grid as a linearly weighted concatenation of multiple neural voxel grids basis. 
The number of the voxel grids basis is the same as the dimension of 3DMM expression basis and the weights of the concatenation is exactly the 3DMM expression coefficients. We define this representation as motion-aware neural voxels. For the canonical appearance, we simply use a single neural voxel grid. Overall, our method can reconstruct a photo-realistic 3D head avatar in 5 minutes implemented by pure Pytorch code. Both quantitative and qualitative experiments demonstrate the superiority of our method in training speed while preserving a comparable rendering result. 

\section{Related Works}

\begin{figure*}[ht]
  \centering
  \includegraphics[width=\linewidth]{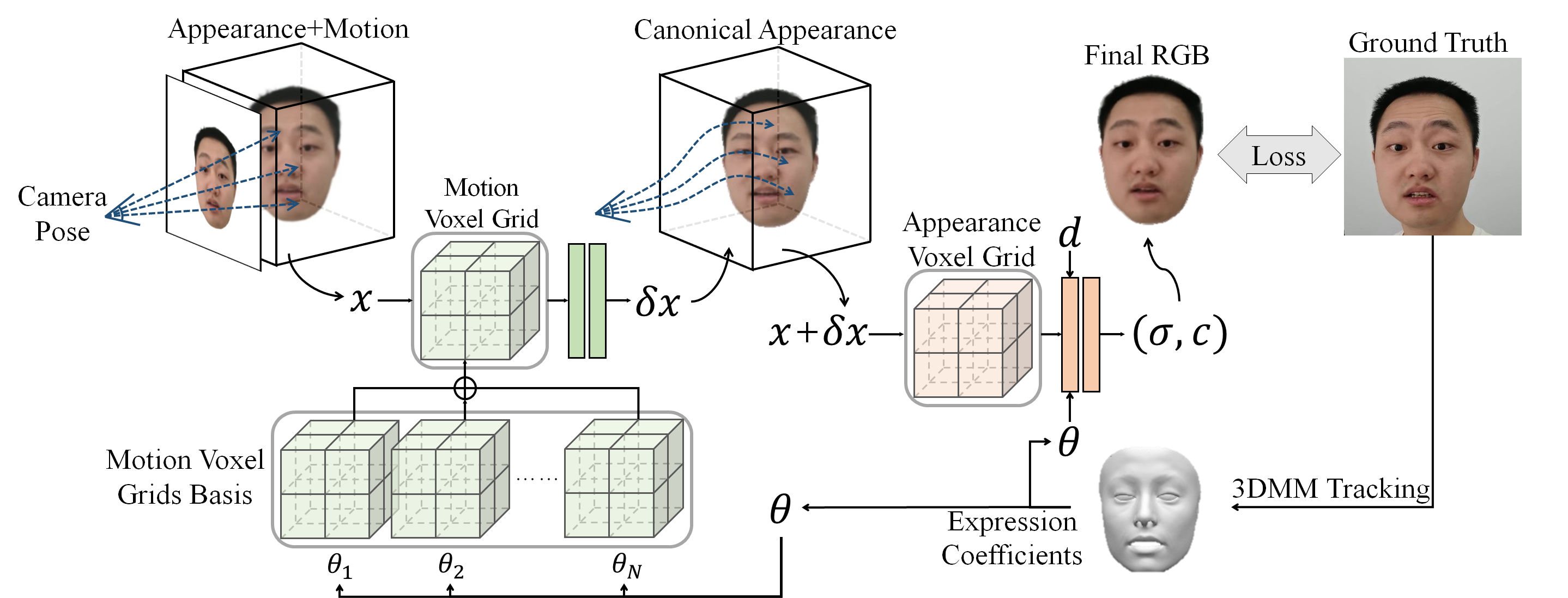}
  \caption{Overview. Given a portrait video, we first track the expression and head pose using a 3DMM template. After the pre-processing, given expression coefficients, we use motion voxel grid basis to represent motions caused by each expression basis and sum them weighted as an entire motion voxel grid. The entire motion voxel grid and the following 2-layer MLP will then transfer an input point $x$ to $x + \delta x$ by adding all expression-related deformations. Finally, we will query point $x + \delta x$ in the appearance voxel grid and generate a final portrait image using volumetric rendering.}
  \label{fig:overview}
\end{figure*}

\textbf{Portrait Video Synthesis.}
In the past period of time, a large number of portrait video synthesis methods have been proposed. The earliest approaches~\cite{weise2011realtime, thies2015real, vlasic2005face, thies2016face2face, li2012a} track to reconstruct textured face mesh with expressions and then re-render it to synthesize portrait images with desired expressions. 
Such methods heavily rely on accurate tracking methods and high-quality reconstructed mesh models etc, which severely limits the broad applications. Warping-based methods~\cite{wiles2018x2face, avebuch2017bringing, nirkin2019fsgan, siarohin2019first, geng2018warp, yin2022styleheat} model the 2D motion flow map as a common representation for expression transfer, but fail to deal with extreme head poses or expressions. Some recent approaches~\cite{drobyshev2022megaportraits, wang2021one} lift the 2D motion flow map into 3D space to overcome such artifacts to a certain extent. Based on the widely-used facial parametric models~\cite{blanz1999a, gerig2018morphable, li2017learning, wang2022faceverse}, template-based methods~\cite{thies2019deferred, koujan2020head2head, doukas2020head2head++, kim2018deep, doukas2021headgan, buehler2021varitex, wang2023styleavatar} that combine face template and neural rendering has gradually become the mainstream. These methods first render coarse images using face templates with controllable expression and pose coefficients, and then utilize convolutional neural networks to generate mouths, hair and rich details to synthesize high-quality portrait images. Other similar methods~\cite{zakharov2019few, chen2020puppeteergan} use semantic maps or landmarks for coarse-level representation to replace face models. 

\textbf{Monocular 3D Head Avatar Reconstruction.} It has always been a challenging task to reconstruct 3D full head avatar from monocular video. 
Early methods~\cite{cao2016real, ichim2015dynamic, hu2017avatar, cao2015real, nagano2018pagan} use blendshape-based templates to fit portraits in input videos to model human heads. For parts such as eyeball and mouth interior that are difficult to model by the templates, these methods require additional post-processing or leverage neural networks to learn geometry or textures~\cite{grassal2022neural, Khakhulin2022realistic}. Inspired by the 3D scene reconstruction method IDR~\cite{yariv2020multiview} based on implicit representation~\cite{park2019deepsdf}, IMAvatar~\cite{zheng2022imavatar} proposes to optimize an occupancy field and a color field to represent the head model based on the FLAME model. As NeRF~\cite{mildenhall2020nerf} representation shows strong ability to synthesis high-fidelity photo-realistic images, Recent methods~\cite{park2021nerfies, park2021hypernerf, shao2023tensor4d} propose to model a 4D scene by a deformable NeRF which can not be animated. As the solution, avatar reconstruction methods~\cite{athar2022rignerf, athar2021flame, gafni2021dynamic, zheng2023avatarrex} use a face template as the condition to learn a controllable NeRF model. Furthermore, NeRFBlendShape~\cite{gao2022reconstructing} use a voxel-based representation to replace the MLP, and model the dynamic NeRF by linear combination of multiple NeRF basis one-to-one corresponding to semantic blendshape coefficients. In the field of the audio-driven avatar, latest methods~\cite{guo2021ad, liu2022semantic} also leverage dynamic NeRF as the representation of head avatars. Besides, some other methods~\cite{hong2022headnerf, zhuang2022mofanerf, wang2022morf, sun2022ide, sun2023next3d, sun2021fenerf, chan2022efficient} are not limited to training person-specific avatars, but to train a general human head model as prior on large-scale datasets. NeRF representation is used in our method as well, but we make great progress in accelerating training speed by both using explicit representation and the decomposition of dynamic motion and static appearance. 


\textbf{Training Acceleration for NeRF.} As vanilla NeRF\cite{mildenhall2020nerf} usually takes hours or even days to complete the training of a static scene, a lot of works focus on speeding up NeRF training process. DVGO~\cite{sun2022direct} proposes to accelerate NeRF training by directly replacing most MLPs with voxel grids, which significantly reduces the time required for training convergence. Plenoxels~\cite{yu_and_fridovichkeil2021plenoxels} goes a step further and proposes to only use a sparse grid with density and spherical harmonic coefficients at each voxel without any neural network. TensoRF~\cite{chen2022tensorf} proposes to decompose the voxel grid into sum of vector-matrix outer products. It reduces the size of the model while ensuring training efficiency and rendering quality. Instant-NGP~\cite{mueller2022instant} introduces multi-resolution hash embedding for the voxel grid structure. Combined with customized CUDA implementation, they enable extremely fast NeRF training. These methods mainly focus on static scenes reconstruction. For dynamic scenes, TiNeuVox~\cite{fang2022fast} proposes to replace the MLPs for canonical scene in D-NeRF~\cite{pumarola2020d} with an explicit voxel grid. Different from~\cite{fang2022fast}  which still represent the time-dependent deformation field with a MLP, our method propose to utilize the explicit data structure for both the static content and the expression-dependent deformation field.
\section{Method}

AvatarMAV can generate a fast 3D head avatar by decoupling complex expression motion from the canonical appearance and voxel-based representation. In this section, we will introduce our voxel-based representation and the training process.

\subsection{Representation}
\label{sec:sec:representation}
Generally, previous methods~\cite{gafni2021dynamic, gao2022reconstructing, guo2021ad, liu2022semantic} formulate NeRF-based head avatar as a expression-dependent NeRF model:
\begin{equation}
 (c, \sigma) = \Phi(x, d, \theta).
\end{equation}
Given a query point $x$ with view direction $d$ and expression coefficients $\theta$, the color and the density denoted by $c$ and $\sigma$ respectively are computed for volumetric rendering~\cite{mildenhall2020nerf}. 

However, their practice of mixing the dynamics expression motion with the appearance of the human head brings obstacles to the fast convergence of avatar learning. 
In contrast, our method decouples the NeRF-based head avatar into the canonical appearance and expression motion based on the physical assumption that a face geometry with expressions can be deformed from the neutral face geometry.
Specifically, we define $\Phi(\cdot)$ as expression-independent NeRF and additionally introduce an expression-dependent deformation $\Omega(\cdot)$, which can be formulated as:
\begin{equation}
 (c, \sigma) = \Phi(x + \delta x, d),
\end{equation}
\begin{equation}
 \text{with} \quad \delta x = \Omega(x, \theta).
 \label{qua:omega}
\end{equation}
Note that $\Phi(\cdot)$ represents a static NeRF and does not contain the expression $\theta$ in its parameters. Since the degrees of freedom of the parameters are greatly reduced, the training speed can be significantly improved as shown in our experiments.

To further improve training efficiency, we enhance our representation with an explicit voxel-based strategy. 
However, using a single voxel grid is infeasible to model the dynamic expression motion.
Previous methods~\cite{pumarola2020d, fang2022fast} utilize deep MLPs to alleviate this issue but lead to a quite time-consuming convergence process.
In our solution, we take the expression prior from 3DMM and use its expression coefficients to condition multiple voxel grids, which is the key to efficient and effective avatar creation.

\textbf{Expression Motion.} 
In the traditional PCA-based 3DMM~\cite{blanz1999a, gerig2018morphable}, the face model can be deformed through a linear combination of its expression PCA basis.
To leverage the expression prior of 3DMM and the expressive power of neural voxels, we use the expression coefficients as the weights to concatenate multiple Motion Voxel Grid (MVG) basis into an entire MVG to represent the expression motion.
Specifically, each dimension of MVG basis uses the same expression coefficient $\theta\in\mathbb{R}^{N}$ as the corresponding 3DMM expression basis, where $N$ denotes the dimension of 3DMM expression basis. From another point of view, MVG basis can be considered as ``neural'' expression basis for head motion, which is able to present more detailed motion in a wider range (including the ears and hair regions). The process can be formulated as:
\begin{equation}
 \bm{V}_{d}(\theta) = \theta^{1}\bm{V}_{d}^{1} \oplus \theta^{2}\bm{V}_{d}^{2} \oplus \cdots \oplus \theta^{N}\bm{V}_{d}^{N},
\end{equation}
where $\{ \bm{V}_{d}^{1}, \bm{V}_{d}^{2},\cdots,\bm{V}_{d}^{N}\}\in \mathbb{R}^{N \times C_{d} \times L_{d} \times L_{d} \times L_{d}}$ denotes the MVG basis in the latent space. $\bm{V}_{d}(\theta) \in \mathbb{R}^{NC_{d} \times L_{d} \times L_{d} \times L_{d}}$ denotes the MVG and $\theta=\{\theta^{1}, \theta^{2},\cdots,\theta^{N}\} \in \mathbb{R}^{N}$ denotes the expression coefficients derived from 3DMM. $C_{d}$ denotes the channel number of voxel features in each voxel grid base. 

Finally, for a query point $x$, we adopt multi-distance interpolation~\cite{fang2022fast} to sample the corresponding feature vector $v_{d}$ and feed it into a 2-layer MLP $f_{d}$ to predict the motion offsets of input points. This process can be formulated as:
\begin{equation}
 v_{d} = T_{K}(x, \bm{V}_{d}(\theta))
\end{equation}
\begin{equation}
 \delta{x} = f_{d}(v_{d}),
\end{equation}
where $T_{K}(\cdot)$ denotes K-distance interpolation~\cite{fang2022fast} and $\delta{x}$ denotes the motion offset of a query point $x$. We have described in detail how we formulate the expression-dependent deformation $\Omega(\cdot)$ in Eq.\ref{qua:omega} using Motion-Aware Neural Voxels representation.

\textbf{Canonical Appearance.} 
Since we model the canonical appearance $\Phi(\cdot)$ as a static NeRF, we just directly represent the canonical appearance field with a single appearance voxel grid $\bm{V}_{a}\in\mathbb{R}^{C_{a} \times L_{a} \times L_{a} \times L_{a}}$ and a two-layer MLP $F_a$, where $C_{a}$ denotes the channel number of voxel features and $L_{a}$ denotes the resolution of the appearance voxel grid. 
The explicit appearance voxel grid $\bm{V}_{a}$ contains information for appearance in its per-voxel features, while the two-layer MLP $F_a$ translates the sampled features to the color and the density. 

Specifically, for each canonical query point $x$, its corresponding feature vector $v_{a}$ is first sampled from the appearance voxel grid $\bm{V}_{a}$ by multi-distance interpolation~\cite{fang2022fast}. 
Then, we feed this feature vector together along with the view direction $d$ into the 2-layer MLP $F_a$ to predict the color and the density:
\begin{equation}
 (c(x), \sigma(x)) = F_{a}(\gamma(v_{a}), \gamma(d), \theta),
\end{equation}
where $c(\cdot)$ and $\sigma(\cdot)$ denote the RGB value and the density value of query point $x$ respectively, $\gamma(\cdot)$ is the positional encoding~\cite{mildenhall2020nerf} mapping the feature vector $v_{a}$ and the view direction $d$ to their periodic formulation.
As it is difficult to model dynamic details and topology changes of the human face such as wrinkles given only motion information, we additionally feed the expression coefficients $\theta$ to $F_a$ to describe the slightly dynamic appearance.

\begin{figure}
  \centering
  \includegraphics[width=1.0\linewidth]{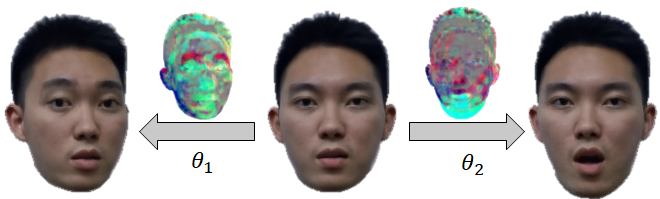}
  \caption{We visualize our learned canonical appearance and expression motion. The color value reflecting the 3D vector of flow}
  \label{fig:decoupling}
\vspace{-0.4cm}
\end{figure}

\subsection{Training}
\label{sec:sec:training}
For more effective training, we first remove the background~\cite{shanchuan2021robust}, the neck, and the body part of the human body~\cite{faceparsing} from each video frame during the data preprocessing phase. Then we detect face landmarks~\cite{openseeface} and obtain expression coefficients and head poses by tracking 3DMM models on each frame. During training, we optimize the voxel grids and MLPs with direct photometric supervision. Moreover, we empirically find that the motion-aware neural voxels may also learn to model the information of canonical appearances such as global constant offset and non-zero offsets in static areas without regularization.
Hence, we add a regularization term to punish all non-zero offsets of sampled points. Meanwhile, as all the offsets in the motion voxel grid are pushed to zeros, the canonical appearance tends to present a neutral expression. The total loss function can be formulated as:
\begin{equation}
\mathcal{L} = \sum_{r \in \mathcal{R}} ||I(r) - I_{gt}(r)||_{1} + \lambda \sum_{r \in \mathcal{R}}\sum_{t \in \mathcal{T}(r)}||\delta(r(t))||_{2},
\end{equation}
where $\mathcal{R}$ denotes the sampled rays in a batch. $I_{gt}$ denotes the preprocessed ground-truth image, $\mathcal{T}(r)$ denotes the distances set of the sampled points on rays $r$, and $\lambda$ denotes the weight of the regular term.

\section{Experiments}

\begin{figure*}[ht]
  \centering
  \includegraphics[width=1.0\linewidth]{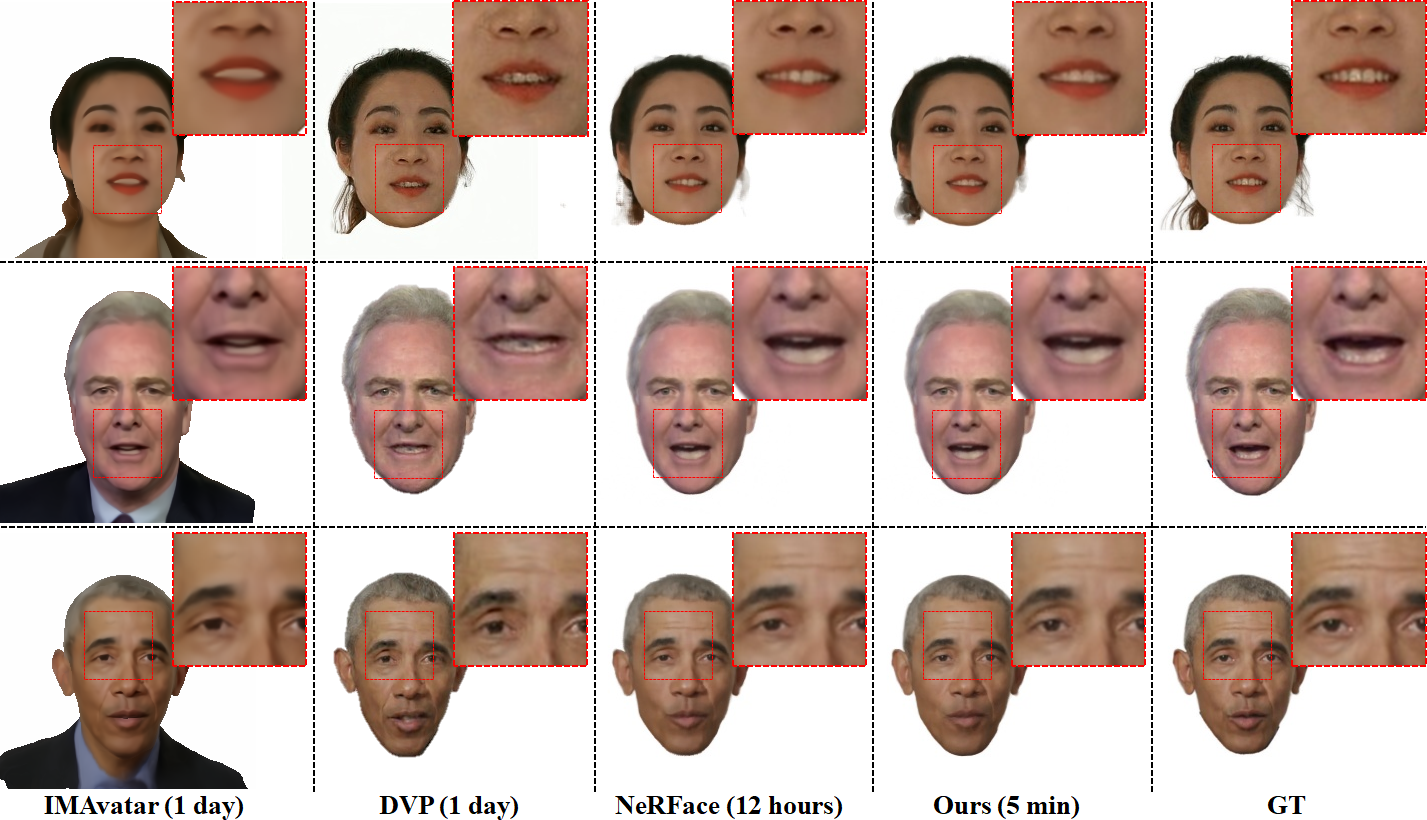}
  \caption{Qualitative comparisons between AvatarMAV and other three state-of-the-art methods on self reenactment task. From left to right: IMAvatar~\cite{zheng2022imavatar}, DVP~\cite{kim2018deep}, NeRFace~\cite{gafni2021dynamic}, AvatarMAV and Ground Truth. Our approach is able to converge and learn details within a short time.}
  \label{fig:self_qualitative}
\end{figure*}

\subsection{Implementation Details}
We implement our whole framework with pure PyTorch. For expression coefficients, we directly use the first 32 expression basis of Basel Face Model~\cite{blanz1999a, gerig2018morphable}. For appearance voxel grid of canonical appearance, we set the channel number of feature as 4 and resolution as $64^{3}$. For expression motion, the number of MVG basis is set as 32 (equal to the number of expression coefficients). The channel number of the features is set as 2 and the resolution of each grid is set as $16^{3}$. All the MLPs in our framework are 2-layer with 64 neurons for each hidden layer. We uniformly set frequency number as 4 for positional encoding. For multi-distance interpolation, we set $K$ as 3. Under this hyperparameter setting, an average of 140ms to render a $256\times256$ resolution image is required on one RTX 3090 GPU.

During optimization, we use Adam optimizer. The initial learning rate is set as $1 \times 10^{-2}$ for MVG basis and appearance voxel grid, and $1 \times 10^{-3}$ for all the tiny MLPs. At the 500 and 2000 iterations, we reduce the learning rate by a third. For the loss function, we set $\lambda=0.01$. For ray sampling, 4096 rays and 64 points are sampled along each ray in each iteration. The batch size is set as 1. Thus, the time consumption of one iteration is 34ms . We train a model for 10000 iterations in total. For the first 6000 iterations, we use training images with $256\times256$ resolution and for the last 4000 iterations, we use $512\times512$ resolution.

We collected 9 training videos for our experiments. with 4 of them are from HDTF dataset~\cite{zhang2021flow}, 1 from NeRFBlendShape~\cite{gao2022reconstructing}. We additionally collect 4 videos by a hand-hold mobile phone. Each video contains about 2000-4000 frames and we reserved the last 15\% of each video solely for evaluation. Meanwhile, we collect several videos as source videos for cross-identity reenactment task.

\begin{figure*}[ht]
  \centering
  \includegraphics[width=1.0\linewidth]{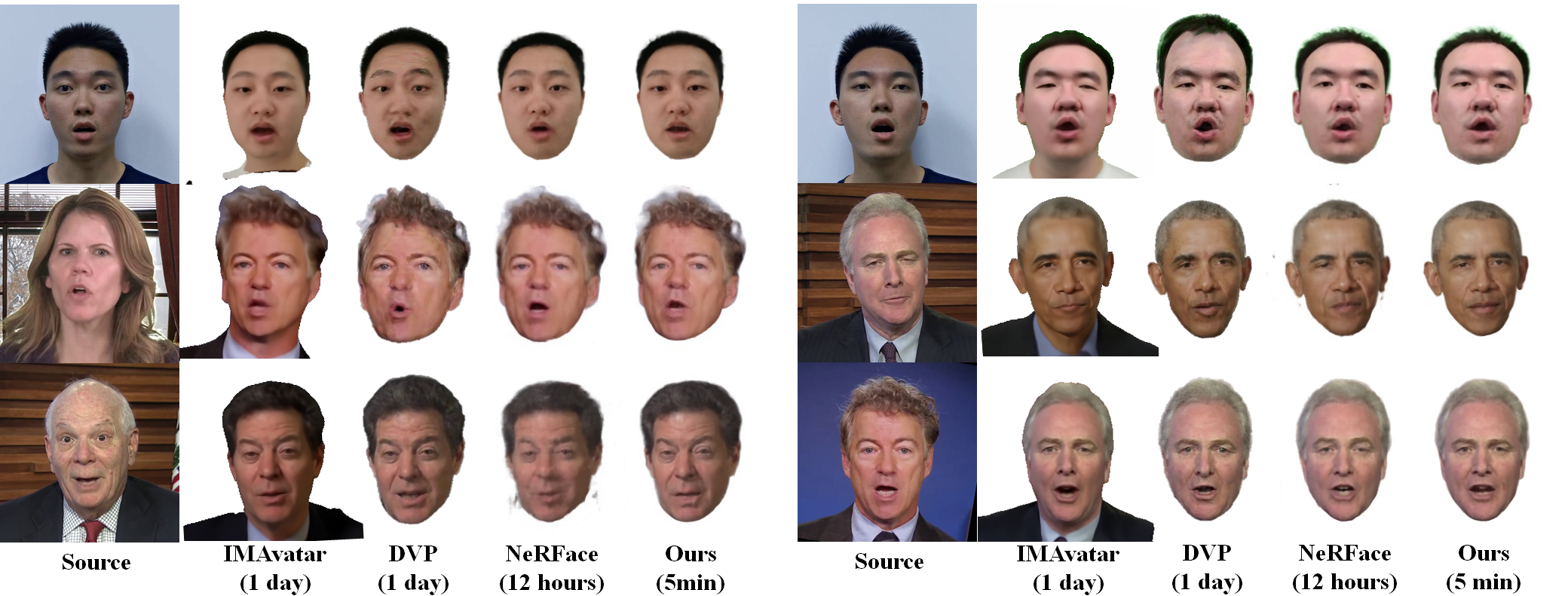}
  \caption{Qualitative results of AvatarMAV and three other state-of-the-art methods on cross-identity reenactment task. From left to right: DVP~\cite{kim2018deep}, IMAvatar~\cite{zheng2022imavatar}, NeRFace~\cite{gafni2021dynamic} and AvatarMAV.}
  \label{fig:facial_reenactment}
  \vspace{-0.0cm}
\end{figure*}

\begin{figure}[ht]
  \centering
  \includegraphics[width=1.0\linewidth]{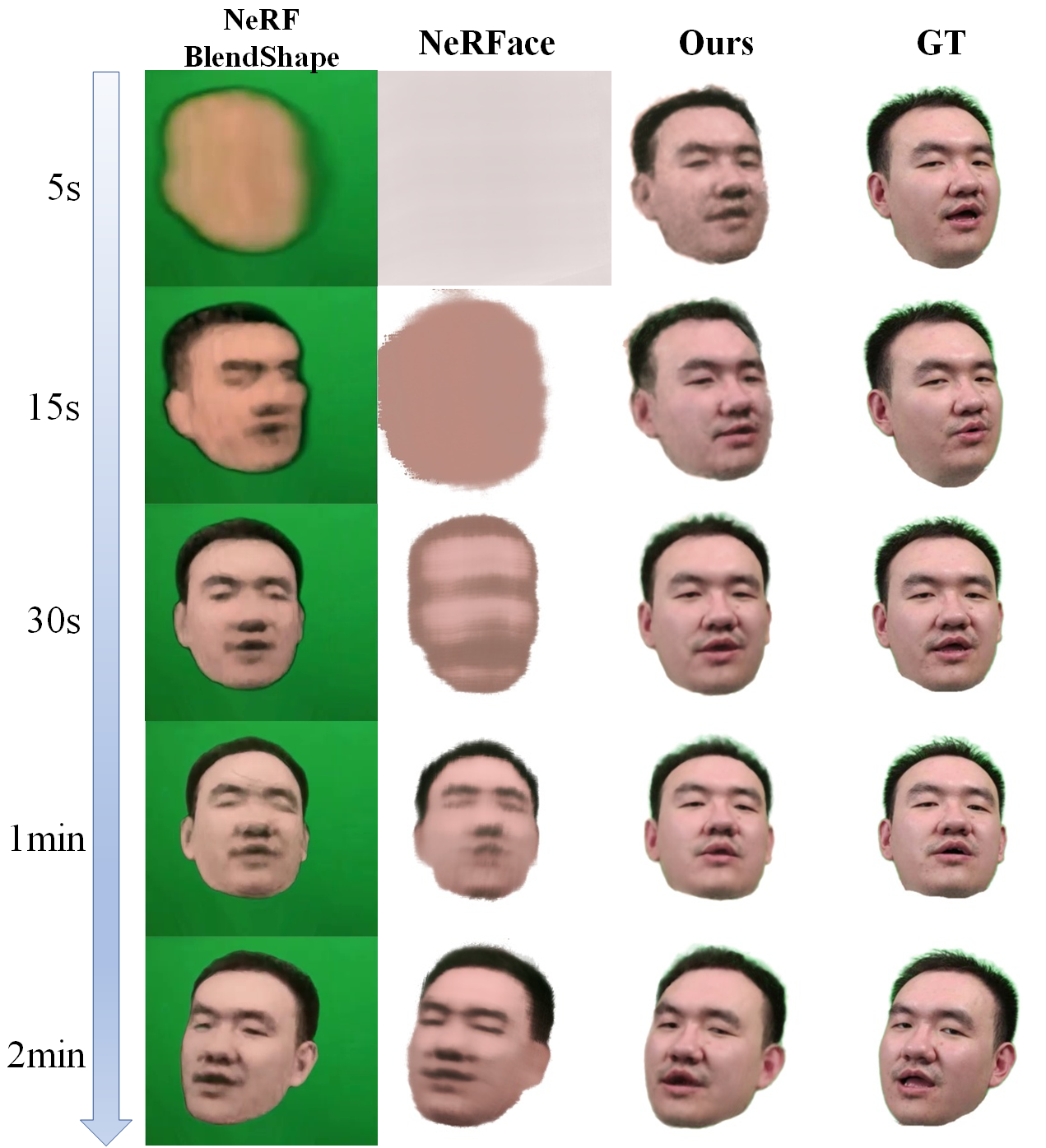}
  \caption{We make qualitative comparisons on training speed among NeRFace~\cite{gafni2021dynamic}, NeRFBlendShape~\cite{gao2022reconstructing} and our AvatarMAV. Our model converges rapidly within the first 2 minutes.}
  \label{fig:comparisons_training}
\vspace{-0.4cm}
\end{figure}

\subsection{Comparisons on Render Quality}
We conduct qualitative and quantitative comparisons on render qualities between our AvatarMAV and three state-of-the-art methods DVP (Deep Video Portraits)~\cite{kim2018deep}, IMAvatar~\cite{zheng2022imavatar}, and NeRFace~\cite{gafni2021dynamic} on both self reenactment task and cross-identity reenactment task. DVP does not reconstruct the full head model, but directly synthesizes 2D images through an image-to-image translation network with a controllable coarse 3DMM model as guidance. IMAvatar reconstructs an implicit signed distance field based on the FLAME model. NeRFace reconstructs a NeRF head model with 3DMM expression coefficients as condition. In our experiments, we spend enough time for training to ensure complete convergence for all the methods. Respectively, \textbf{1 day} for IMAvatar~\cite{zheng2022imavatar}, \textbf{1 day} for DVP~\cite{kim2018deep}, \textbf{12 hours} for NeRFace~\cite{gafni2021dynamic}, and \textbf{5 minutes} for AvatarMAV. Since the concurrent work on NeRFBlendShape~\cite{gao2022reconstructing} doesn't release their training code, we only make qualitative comparisons on training speed to their demo video in the next section. 

Qualitative comparisons on self reenactment task are shown in Fig.~\ref{fig:self_qualitative}. The results validate that AvatarMAV achieves the highest render quality while the training time is far less than the other methods. IMAvatar reconstructs an implicit model based on the mesh of a FLAME template, yet the expressiveness is limited. DVP inherits the GAN framework to generate images. But in many cases, the generated details are not appropriate. The performance of NeRFace is comparable to AvatarMAV, but the training time is much longer.

\begin{table}[ht]
\centering
\caption{Quantitative evaluation results of AvatarMAV and other three state-of-the-art methods on self reenactment task. We calculate mean values on three different subjects.}
\begin{tabular}{c|c|c|c|c}
\hline
Method                              & MSE $\downarrow$     & PSNR $\uparrow$    & SSIM                & LPIPS                 \\
\hline
DVP~\cite{kim2018deep}              & 0.0047               & 24.2               & 0.89                & 0.072                 \\
IMavatar~\cite{zheng2022imavatar}   & 0.0041               & 24.6               & 0.92                & 0.131                 \\
NeRFace~\cite{gafni2021dynamic}     & 0.0015               & 30.2               & \textbf{0.96}       & \textbf{0.038}        \\
AvatarMAV                           & \textbf{0.0014}      & \textbf{30.4}      & \textbf{0.96}       & \textbf{0.038}        \\
\hline
\end{tabular} 
\label{tab:self_quantitative}
\vspace{-0.1cm}
\end{table}

Table.~\ref{tab:self_quantitative} shows the quantitative evaluation results. We evaluate on four metrics: Mean Squared Error (MSE), Peak Signal-to-Noise Ratio (PSNR), Structure Similarity Index (SSIM) and Learned Perceptual Image Patch Similarity (LPIPS). Our AvatarMAV achieves the best results in MSE and PSNR metrics, and comparable results to NeRFace on both SSIM and LPIPS metrics.

Next, we compare AvatarMAV with these three state-of-the-art methods on cross-identity reenactment task. Qualitative results are shown in Fig.~\ref{fig:facial_reenactment}. In cases where the expression from the source video is out of the distribution in the training data, NeRFace might produce floating artifacts. Benefiting from our decoupling of the dynamic and the static elements, AvatarMAV maintains better stability.

\subsection{Comparisons on Training Speed} 
We qualitatively compare the training speed of our AvatarMAV and two other NeRF-based methods: NeRFBlendShape~\cite{gao2022reconstructing} and NeRFace~\cite{gafni2021dynamic}.  We train each model from scratch, and render the corresponding image at 5s, 15s, 30s, 1min, 2min. Since our model has almost reached complete convergence within the first 2 minutes. We do not visualize the render results after 2 minutes. NeRFBlendshape claims that their method takes 20 minutes to converge. In our experiments, we found that NeRFace actually only takes a few hours to converge. Note that, We use the subject in the demo video of NeRFBlendShape for comparison.

Qualitatively comparisons are shown in Fig.~\ref{fig:qualitative_ablation}. Our model converges with very high efficiency. In the first 30 seconds of training, our model can complete most of the convergence after very few iterations. At the time of 2 minutes, our model has almost converged (5 minutes for the complete convergence). In contrast, much more training time is required for the other two methods.
\section{Ablation Study}

\subsection{Complete Voxel-based Representation}

\begin{figure}[ht]
  \centering
  \includegraphics[width=1.0\linewidth]{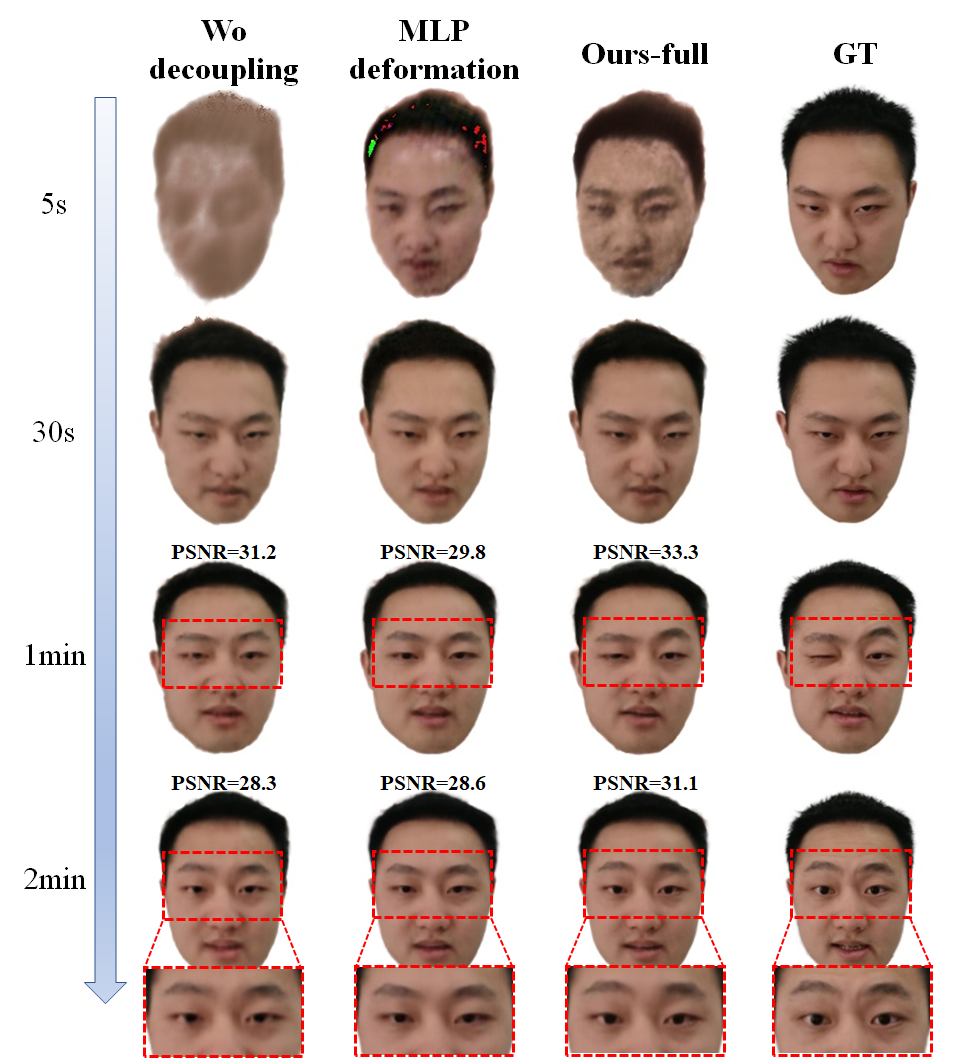}
  \caption{Qualitative results of two baselines and our AvatarMAV on self reenactment task at different points in time. Our full method AvatarMAV can learn both high-quality appearance and expression motion faster.}
  \label{fig:qualitative_ablation}
\vspace{-0.0cm}
\end{figure}

To accelerate training, our idea consists of two key points in improving the representation of NeRF-based head avatar: decoupling expression motion from canonical appearance; modeling the expression motion by motion-aware neural voxels representation. In the following, we ablate these two components.

\textbf{Without Decoupling.}
This baseline no longer decouples expression motion and canonical appearance. Given $N$ dimension expression coefficients, we also establish $N$ corresponding voxel grids basis but the weighted concatenation of the voxel grids directly represent the final expressive head avatar. A 2-layer percepron is followed to predict the color and the density values. In our expriments, we set the resolution of the voxel grid basis as $64^{3}$, which is equal to resolution of the appearance voxel grid in our full method.

\textbf{MLP Deformation.}
In this baseline, we directly use a MLP to implicitly model the expression motion. Specifically, for a query point, we feed the coordinate concatenated with the expression coefficients in to a MLP with positional encoding to obtain the offset. In our expriments, we use a 4-layer percepron with 128 neurons for each hidden layer. For the canonical appearance, we still use a neural voxel grid with a 2-layer percepron.

In this section, we evaluate our full method AvatarMAV and the two variant baselines above. 
Qualitative and quantitative results are shown in Fig.~\ref{fig:qualitative_ablation} and Table.~\ref{tab:quantitative_ablation_latentvoxel}. As the without decoupling baseline does not decouple the expression motion from the canonical appearance, the rendered images tend to be blurred. the MLP deformation baseline benefits from decoupling and converges faster. However, it can be observed from the images at 1min or 2min in Fig.~\ref{fig:qualitative_ablation} that only voxel-based canonical appearance converges to the average state rapidly, while MLP-based motion converges slowly. Our AvatarMAV combines the advantages of both decoupling and voxel-based representation to achieve the best training efficiency. We also mark the corresponding quantitative evaluation results (PSNR) of the two frames, which also verified the above conclusions.

\begin{table}[ht]
\centering
\caption{Quantitative evaluation results of all the ablation baselines and AvatarMAV on self reenactment task at 30 seconds and 5 minutes. We calculate mean values on three different subjects.}
\begin{tabular}{c|c|c|c|c}
\hline
Method (\textbf{30s})          & MSE                  & PSNR               & SSIM                & LPIPS                 \\
\hline
Wo decoupling                  & 0.0032               & 26.0               & 0.926               & 0.073                 \\
MLP deformation                & 0.0037               & 25.4               & 0.919               & 0.070                 \\
Direct motion voxels           & 0.0030               & 26.4               & 0.931               & 0.061                 \\
AvatarMAV                      & \textbf{0.0025}      & \textbf{27.1}      & \textbf{0.938}      & \textbf{0.062}        \\
\hline
\hline
Method (\textbf{5min})         & MSE                  & PSNR               & SSIM                & LPIPS                 \\
\hline
Wo decoupling                  & 0.0021               & 28.0               & 0.946               & 0.062                 \\
MLP deformation                & 0.0028               & 26.7               & 0.933               & 0.065                 \\
Direct motion voxels           & 0.0023               & 27.9               & 0.945               & 0.053                 \\
AvatarMAV                      & \textbf{0.0020}      & \textbf{28.5}      & \textbf{0.950}      & \textbf{0.052}        \\
\hline
\end{tabular} 

\label{tab:quantitative_ablation_latentvoxel}
\vspace{-0.1cm}
\end{table}

\subsection{Latent Voxel}

\begin{figure}[ht]
  \centering
  \includegraphics[width=1.0\linewidth]{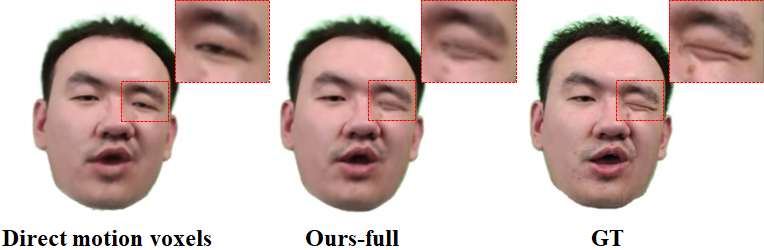}
  \caption{Qualitative results of direct motion voxels and AvatarMAV on self reenactment task (5min). AvatarMAV can learn finer expressions.}
  \label{fig:qualitative_ablation_latentvoxel}
\vspace{-0.3cm}
\end{figure}

An alternative idea is to directly store offset value $\delta x$ rather than feature vector in the described MVG basis in Sec.~\ref{sec:sec:representation}, such that the tiny MLP $f_d$ followed can be removed to achieve higher efficiency. We call the architecture direct motion voxels to distinguish it from the neural motion voxels. Meanwhile, the MVG is calculated by weighted blending of MVG basis to match the subsequent process. We evaluate this variant baseline qualitatively and quantitatively as shown in Fig.~\ref{fig:qualitative_ablation_latentvoxel} and Table.~\ref{tab:quantitative_ablation_latentvoxel}. As a result, the direct motion voxels leads to a decrease in the performance. To explain, the motion field in AvatarMAV is a backward warp field that maps the points in the live space to the canonical space. Such a backward warp field is ambiguous as there exist non-linear many-to-one mappings. To handle this issue, an MLP is necessary to achieve such non-linear mapping.

\section{Limitation}

\begin{figure}[ht]
  \centering
  \includegraphics[width=0.9\linewidth]{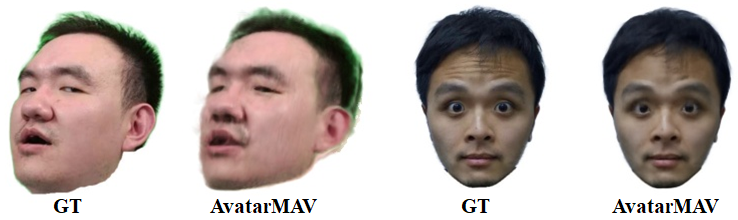}
  \caption{Failure cases when the rendering viewpoint is away from the front view (left) or the expression is exaggerated (right).}
  \label{fig:failure_case}
  \vspace{-0.3cm}
\end{figure}

Although we accelerate the training speed of NeRF-based head avatar to almost instant, there is no significant improvement in rendering quality. As the camera poses and expression coefficients obtained by fitting a face template are not accurate enough. Artifacts and blurriness appear when the rendering viewpoint is away from the front view or the expression is exaggerated as shown in Fig.~\ref{fig:failure_case}. On the other hand, limited by the face template, our approach can only handle human heads, without the ability to reconstruct other regions such as neck and upper body. In the future, we will try to use more general parameterized representation to solve this problem.

\section{Conclusion}
\label{sec:Conclusion}
In this paper, we have presented AvatarMAV, a fast 3D head avatar reconstruction method using motion-aware neural voxels. The proposed neural-voxel-based representation of both the canonical appearance and the motion field is able to greatly accelerate the training process. Furthermore, we have also demonstrated superiority and efficiency of the proposed motion-aware neural voxels in reconstructing complex expression-related motion. In future work, on-line creating a 3D head avatar during live capturing process might become feasible, which can greatly reduce the cost of generating digital human faces. We believe AvatarMAV will inspire the following facial reenactment researches, and the proposed motion-aware neural voxels could further broaden the applications of dynamic NeRF.

\begin{acks}
This paper is supported by National Key R\&D Program of China (2022YFF0902200), the NSFC project No.62125107 and No.61827805.
\end{acks}

\bibliographystyle{ACM-Reference-Format}
\bibliography{base}

\end{document}